\documentclass[10pt,twocolumn,letterpaper]{article}

\usepackage{cvpr}
\usepackage{times}
\usepackage{epsfig}
\usepackage{graphicx}
\usepackage{amsmath}
\usepackage{amssymb}
\usepackage{booktabs} 
\usepackage{url}
\usepackage{algorithm2e}
\usepackage{caption}
\usepackage{mathtools}

\usepackage[symbol]{footmisc}
\usepackage{amsmath}

\usepackage[pagebackref=true,breaklinks=true,letterpaper=true,colorlinks,bookmarks=false]{hyperref}

\cvprfinalcopy 

\def\cvprPaperID{****} 
\def\httilde{\mbox{\tt\raisebox{-.5ex}{\symbol{126}}}}

\ifcvprfinal\fi
\begin{document}


\renewcommand{\thefootnote}{\fnsymbol{footnote}}
\title{Fighting Quantization Bias With Bias}
\author{Alexander Finkelstein\thanks{Equal contribution}
\and
Uri Almog\footnote[1]{Equal contribution}\\
Hailo Technologies\\
{\tt\small alex.finkelstein,uri.almog,mark@hailotech.com}
\and
Mark Grobman
}

\maketitle
\renewcommand*{\thefootnote}{\arabic{footnote}}
\begin{abstract}
   Low-precision representation of deep neural networks (DNNs) is critical for efficient deployment of deep learning application on embedded platforms, however, converting the network to low precision degrades its performance. Crucially, networks that are designed for embedded applications usually suffer from increased degradation since they have less redundancy. This is most evident for the ubiquitous MobileNet architecture \cite{2017arXiv170404861H,2018arXiv180104381S} which requires a costly quantization-aware training cycle to achieve acceptable performance when quantized to 8-bits. In this paper, we trace the source of the degradation in MobileNets to a shift in the mean activation value. This shift is caused by an inherent bias in the quantization process which builds up across layers, shifting all network statistics away from the learned distribution.  We show that this phenomenon happens in other architectures as well. We propose a simple remedy - compensating for the quantization induced shift by adding a constant to the additive bias term of each channel. We develop two simple methods for estimating the correction constants - one using iterative evaluation of the quantized network and one where the constants are set using a short training phase. Both methods are fast and require only a small amount of unlabeled data, making them appealing for rapid deployment of neural networks. Using the above methods we are able to match the performance of training-based quantization of MobileNets at a fraction of the cost.  
   
\end{abstract}

\section{Introduction}
In the last years, an increasing amount of effort is invested into executing DNN inference in low-precision arithmetic.
While very effective in cutting down on memory, compute and power usage, this comes at a price of degraded network performance caused by weight- and activation-rounding errors. Quantization, the conversion of a net to its low-precision version, is performed according to a "scheme", a conceptual model of the hardware execution environment.  In this paper we follow the 8-bit integer quantization scheme used in \cite{Jacob2017} which is widely employed across both cloud- and edge-devices due to its easy and efficient implementation on hardware. For most network architectures, 8-bit quantization carries a minimal degradation penalty \cite{KrishnaGoogle} leading recent research to double-down on more aggressive schemes, using 4 bits or less for activations, weights or both \cite{IBMFAQ2018discover4bit,Soudry4bit, choi2018pact_ReluAlphaTrained, Jung2018}. Nonetheless, some architectures, such as Mobilenet \cite{2017arXiv170404861H,2018arXiv180104381S}, Inception-V1 \cite{Szegedy2014} and Densenet \cite{Huang2016} still exhibit significant degradation when quantized to 8 bits. Mobilenets are of particular interest as they were designed specifically for embedded applications and as such are often deployed in their quantized form. In order to successfully deploy these networks a quantization-aware training phase \cite{QuallcomMobilenet, KrishnaGoogle, Jacob2017} is usually introduced to regain the network accuracy.  This phase is costly in time and requires access to the original dataset on which the network was trained. Moreover, while effective, the training offers little insight into why there was significant degradation in the first place. Recently, there has been more focus on methods that do not rely on an additional training phase \cite{SameSame, Soudry4bit, zhao2019ChannelSplitting, Migacz2017} but an exhaustive 8-bit quantization scheme that excludes training remains an open challenge. 
\newline\indent In this paper, we begin by showing that some networks exhibit a significant shift in the mean activation value following quantization, which we henceforth denote as MAS (mean activation shift). We note that in order for such a shift to have an impact on the performance of the network it must be of significant magnitude with respect to the mean activation itself. The shift is the result of quantization rounding errors being highly unbalanced - a 'small numbers effect' that is statistically implausible for layers with many parameters, but can become the main error source for layers with a small amount of parameters (e.g. depthwise convolution with 9 parameters per channel). Note that the frugal use of parameters is especially prevalent in network architectures aimed at embedded applications \cite{2017arXiv170404861H,2018arXiv180104381S,Zhang2017} making them more susceptible to quantization induced shifts. 
We then show that the shift introduced by the quantization can be canceled by using the bias parameter of the channels. Since the bias term is additive, any constant added to it will shift the activation distribution of the channel. Once the task is defined as fixing the shifts, the problem then becomes how to estimate them. To this end, we develop two procedures - one based on direct estimation over a set of test images and the other performing a fine-tuning phase involving only the bias terms of the network.
The main contributions of this paper are: 
\begin{itemize}
	\setlength\itemsep{1em}
	\item \textbf{Mean activation shift (MAS)}: We explore both the statistical origins and the impact of MAS on layer-level quantization errors. We show that MAS can arise and be responsible for a large component of the error when quantizing network architectures relying on layers with a very small number of parameters (e.g. MobileNet's depthwise layers).  
	\item \textbf{Shift compensation using bias terms}: We show MAS can be compensated by using the bias terms of the network layers, and this alone can drastically reduce degradation, further establishing the previous claim. We propose two algorithms to that effect - Iterative Bias Correction (IBC) and Bias Fine Tune (BFT) - and we analyze their performance on a variety of Imagenet trained classifiers. 
\end{itemize}
While our experiments use convolutional neural networks (CNNs) for classification, we expect both the techniques and the analysis to extend into other tasks (e.g. Object-Detection) building on these classifiers as their feature extractors.
\section{Previous Work}
    
\subsection{Quantization Procedures}
A slew of ideas on DNN quantization were published in recent years, with widely and subtly diverging assumptions about the capabilities of the underlying deployment hardware . For instance, support for non-uniform quantization \cite{xilinxSmallFloat,baskin2018uniq, SongHanDeepCompression}, fine-grained mixed precision \cite{park2018value_mixedprec} or per-channel ("\textit{channelwise}") quantization \cite{KrishnaGoogle, Soudry4bit, choukrounHuawei2019low, goncharenko2018fast}. Leaving the batch-normalization layer unfolded \cite{IBMFAQ2018discover4bit} is another nuance since, similarly to \textit{channelwise quantization}, it enables per-channel scaling at the cost of additional computations. In this work, we restrict ourselves to the simplest setting \textit{folded-batch normalization, layerwise, UINT8-activations, INT8-weights} that was introduced by \cite{Jacob2017}. The attractiveness of this simple setting is that it is supported \textit{efficiently} by most existing hardware on which deep-learning applications are deployed.
The literature can also be dissected by the cost and complexity of the quantization procedures, falling into two main categories:

\paragraph{Post-training quantization:} these methods work directly on a pre-trained network without an additional training phase to compensate for the effects of quantization. Generally lean, these methods have minimal requirements in terms of data, expertise and effort (of both man and machine). Most works are focused on how to best estimate the dynamic range of a layer given its activation distribution. These range from simple heuristics for outlier removal  \cite{KrishnaGoogle, IBMFAQ2018discover4bit}, to more advanced methods employing statistical analysis \cite{Migacz2017, PolinoDistillation,choi2018pact_ReluAlphaTrained, Soudry4bit, choukrounHuawei2019low, goncharenko2018fast}. In this work we use naive min/max statistics (without clipping) to determine dynamic range in order to keep the comparative experiments on IBC/BFT as clean as possible, however, it can be combined with more advanced methods. Other approaches try to lower the dynamic range of the layer by reducing the intra-layer channel range. In \cite{zhao2019ChannelSplitting} it was proposed to improve layerwise quantization by splitting channels with outliers, improving quantization at the cost of additional computations. Finally, in \cite{SameSame} it was proposed to employ inversely-proportional factorization to equalize the dynamic ranges of channels within a layer. 

\paragraph{Quantization-aware training} involves optimizing the quantized weights using a large training set, with net-specific hyperparameters \cite{IBMFAQ2018discover4bit} usually taking the pre-trained net as a starting point. The optimization target may use the ground-truth labels \cite{Jacob2017} as in the normal training, and/or an adaptation of knowledge distillation \cite{hinton2015distilling} loss, namely using the full-precision net as the target ("Teacher"), with the quantized ("Student") net punished for distance of its output to the Teacher's \cite{PolinoDistillation, Apprentice}. Rendering the training "quantization-aware" is non-trivial since the rounding and clipping operations are non-differentiable. The simplest and standard approach is that of a Straight-Through-Estimator (STE) \cite{Bengio_STE, PolinoDistillation} which essentially consists of using the quantized version of the net on forward-pass while using the full-precision network on the backward-pass. One of the methods developed in this work (BFT) employs knowledge-distillation and a STE within a very short training procedure (micro-training) restricted to the network biases only. Concurrently to our work, in \cite{choukrounHuawei2019low} a micro-training of scaling factors (multiplicative rather than additive as in ours) was proposed. Like us, they categorize this as post-training quantization since very little data and training time is used.

\subsection{MobileNets Quantization}
Mobilenets \cite{2017arXiv170404861H, 2018arXiv180104381S} are CNNs based on "separable" convolutions - intermittent depthwise and pointwise convolution layers. Their compact and low-redundancy design makes quantization challenging as they have less resilience to noise. The basic \textit{layerwise} 8-bit quantization of Mobilenet-v1 leads to complete loss of accuracy \cite{QuallcomMobilenet, KrishnaGoogle}. A major source of the degradation is due to issues caused by the folding of batch-normalization to degenerate channels (i.e. channel with constant zero output) \cite{QuallcomMobilenet}. Zeroing out these channels reduces degradation to ~$~10\%$ and we employ this technique as well. Quantization-aware training methods \cite{Jacob2017, KrishnaGoogle} are able to reduce degradation to $~1\%$ for both Mobilenet-v1 and v2. Current \textit{post-training quantization} methods can achieve similarly good results (i.e. $<~1\%$) only by using channelwise quantization \cite{KrishnaGoogle} which is not supported on all hardware. Authors of \cite{QuallcomMobilenet} achieve $~2.5\%$ degradation in a layerwise post-training quantization setting but resort to remodeling the network architecture into a "quantization-friendly" version by changing the activation function and modifying the separable-convolution building block. In \cite{SameSame} a $~3\%$ degradation is achieved without either remodelling or retraining by using inversely-proportional factorization to equalize the dynamic range of channels; this further reduces to $~0.6\%$ when the BFT method described in this paper is applied on top, setting the new state-of-the-art for Mobilenet-v1 and v2 "basic-gemmlowp" quantization while being easy and fast to run. 

\section{Problem Statement and Analysis}\label{sec:Problem Statement}
\begin{figure}[ht]
		\vskip 0.2in
		\begin{center}
			\centerline{\includegraphics[width=\columnwidth]{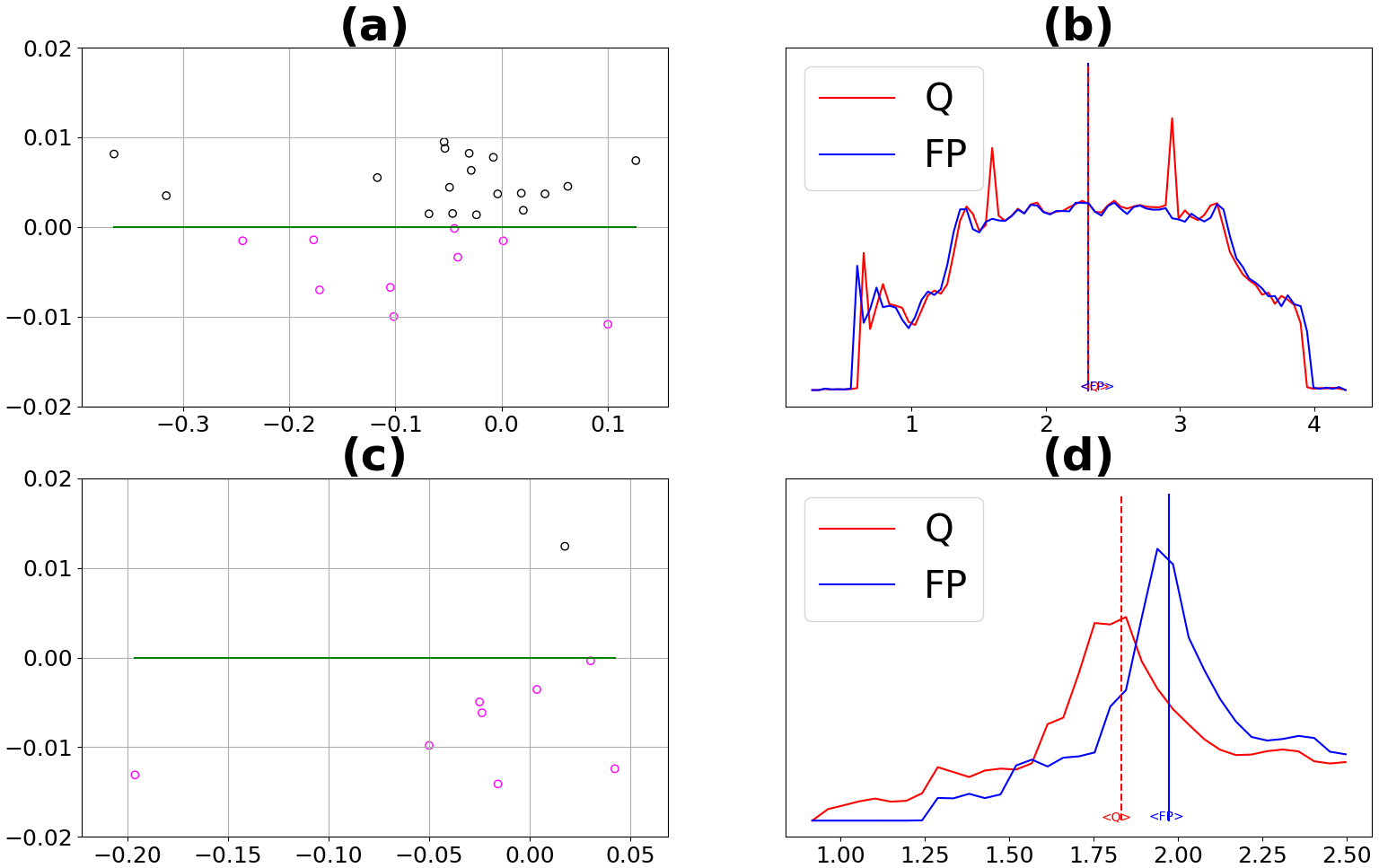}}			
			\caption{Mean activation shift in two different layers of Mobilenet 1 1.0 224: \textbf{Left:} weight rounding errors (vertical axis) vs. full precision (FP) weights (horizontal axis). An ideal transformation would place all the dots on the null horizontal line. Black indicates a weight whose quantized value is greater than the FP, and magenta indicates weights whose quantized value is smaller or equal to the FP value. \textbf{Right:} 32-image activation histograms for the channel corresponding to the weights on the left. \textbf{Top} row is for the \textit{conv1/channel1} kernel slice, with 27 weight elements (\textbf{1.a}) and a small (0.029) relative mean kernel shift, producing a small MAS (\textbf{1.b}). \textbf{Bottom} row is for the are for \textit{depthwise1} layer (\textit{channel12}), with 9 weight elements (\textbf{1.c}). The relative mean kernel error is large, (0.239), resulting in a large MAS. (\textbf{1.d}). The vertical blue and red lines mark the mean activation values for FP and quantized networks, with a clear shift in \textbf{1.d} and an indiscernible shift in \textbf{1.b} }
		\label{fig:DistributionShift}
		\end{center}
		\vskip -0.2in
\end{figure}

A general sentiment underlying quantization of neural networks is that \cite{PeteWarden17}: \begin{quote} "...as long as the errors generally cancel each other out, they'll just appear as the kind of random noise that the network is trained to cope with and so not destroy the overall accuracy by introducing a bias..."\end{quote}
Contrary to the above assumption, in this work we show that when comparing the activations of a feature map in a full-precision and quantized net (QNN), a significant "DC" shift between the distributions can be observed (Fig. \ref{fig:DistributionShift}).
This conceptual gap could arise from thinking on the level of a single tensor, whose rounding error distribution is assumed to be uniform and symmetric; then assuming this property to propagate through the net. However, the law of large numbers doesn't always hold, and small asymmetries might get amplified and accumulate across layers.

In the rest of this section, we first (\ref{subsection3.1}) define the quantization error and it's "DC" component (denoted as MAS). Then in \ref{subsection3.2} we quantify in precise terms the MAS contribution to the error, empirically demonstrating its significance. Finally, in \ref{subsection3.3} we explore further the statistical mechanism of MAS generation and its dependence on layer structure.

\subsection{Quantization Error Measures}\label{subsection3.1}
For a given layer $l$ and feature $ch$ we define the quantization error as the element-wise difference between activations in the original DNN ($x$) and those in the quantized DNN (QNN) ($x^{(q)}$) :
\begin{equation}\label{define_error}
e_{l, ch} = (x^{(q)}_{l, ch}-x_{l, ch}).
\end{equation}
Where $e_{l, ch}, x^{(q)}_{l, ch},x_{l, ch}$ are vectors (for brevity we sometimes drop the subscripts where they are understood from context). Some works make a distinction between rounding "noise" and clipping "distortion"; in this work both are treated indivisibly and we use the terms "noise" and "error" interchangeably. For every channel we define the signal energy, $E(x^2)$, as:
\begin{equation}\label{define_signal_energy}
E(x^2) = \frac{1}{N} \sum{x^2_{l,ch}}
\end{equation}
The expectations $E()$ here and below are to be understood as taken across all pixels of the testing set.
We use the mean rather than sum (as in standard $L_2$ norm definition) so that our analysis will be invariant to the number of elements within a layer. Similarly, we define $E(e^2)$ as the energy of the quantization error. Next, we define the \textbf{Mean Activation Shift} (MAS) as:
\begin{equation}\label{define_shift}
\Delta_{l, ch} = E(x^{(q)}_{l, ch}) - E(x_{l, ch}) = E(x^{(q)}_{l, ch} - x_{l, ch}) = E(e_{l, ch})
\end{equation}

For convenience of exposition we also define we also define the following two quantites: inverse root quantization to noise ratio (rQNSR) and Mean activation Shift to Signal Ratio (MSSR) as:
\begin{equation}\label{define_QNSR}
rQNSR_{l, ch} = \sqrt{\frac{E(e^2)} {E(x^2)}}
\end{equation}
\begin{equation}\label{define_MSSR}
MSSR_{l, ch} = \frac{E(e_{l, ch})} {\sqrt{E(x_{l,ch}^2)}}
\end{equation}
MSSR measures how significant the MAS is compared to the average activation. The rQSNR is a measure of the overall noise level in a channel. Note that MSSR and rQNSR scale similarly, allowing for easy comparison.

All definitions above being random variables' expectations, in practice we use estimates generated by sampling the activations of the network across inferences on a batch of images.

\subsection{The Effect of Mean Activation Shifts}\label{subsection3.2}
\begin{figure}[ht]
		\vskip 0.2in
		\begin{center}
			\centerline{\includegraphics[width=\columnwidth]{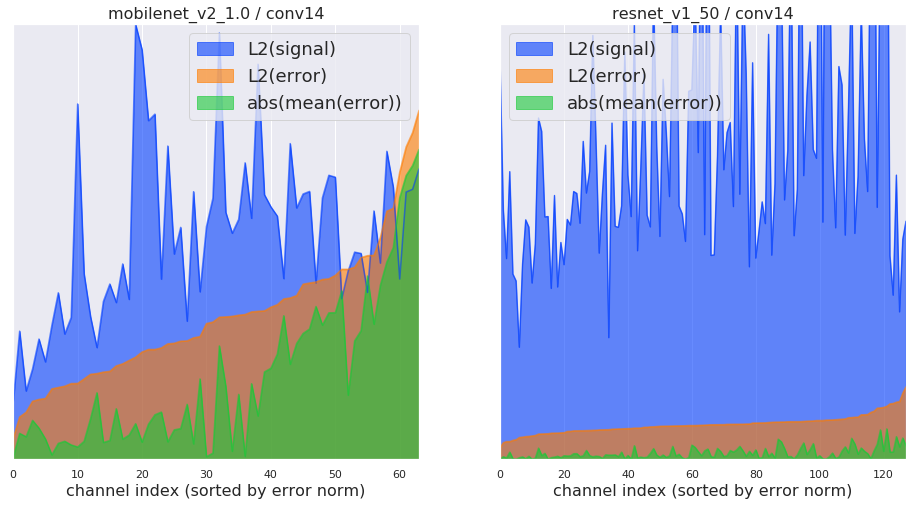}}			
			\caption{32-image estimates of \textbf{(a}) $E(e_{l, ch})$ (the MAS) and \textbf{(b),(c)} $\sqrt{E(x^2_{l, ch})}$, $\sqrt{E(e^2_{l, ch})}$ (L2 norms of activation signal, and its quantization error), per channel, for typical layers of two nets strongly differing in quantization-friendliness. When quantization error is large, the mean-shift tends to be larger even in relative terms, becoming the major contribution to q-error.
			}
		\label{fig:MeanShiftContribution}
		\end{center}
		\vskip -0.2in
\end{figure}

\begin{figure}[ht]
		\vskip 0.2in
		\begin{center}
			\centerline{\includegraphics[width=\columnwidth]{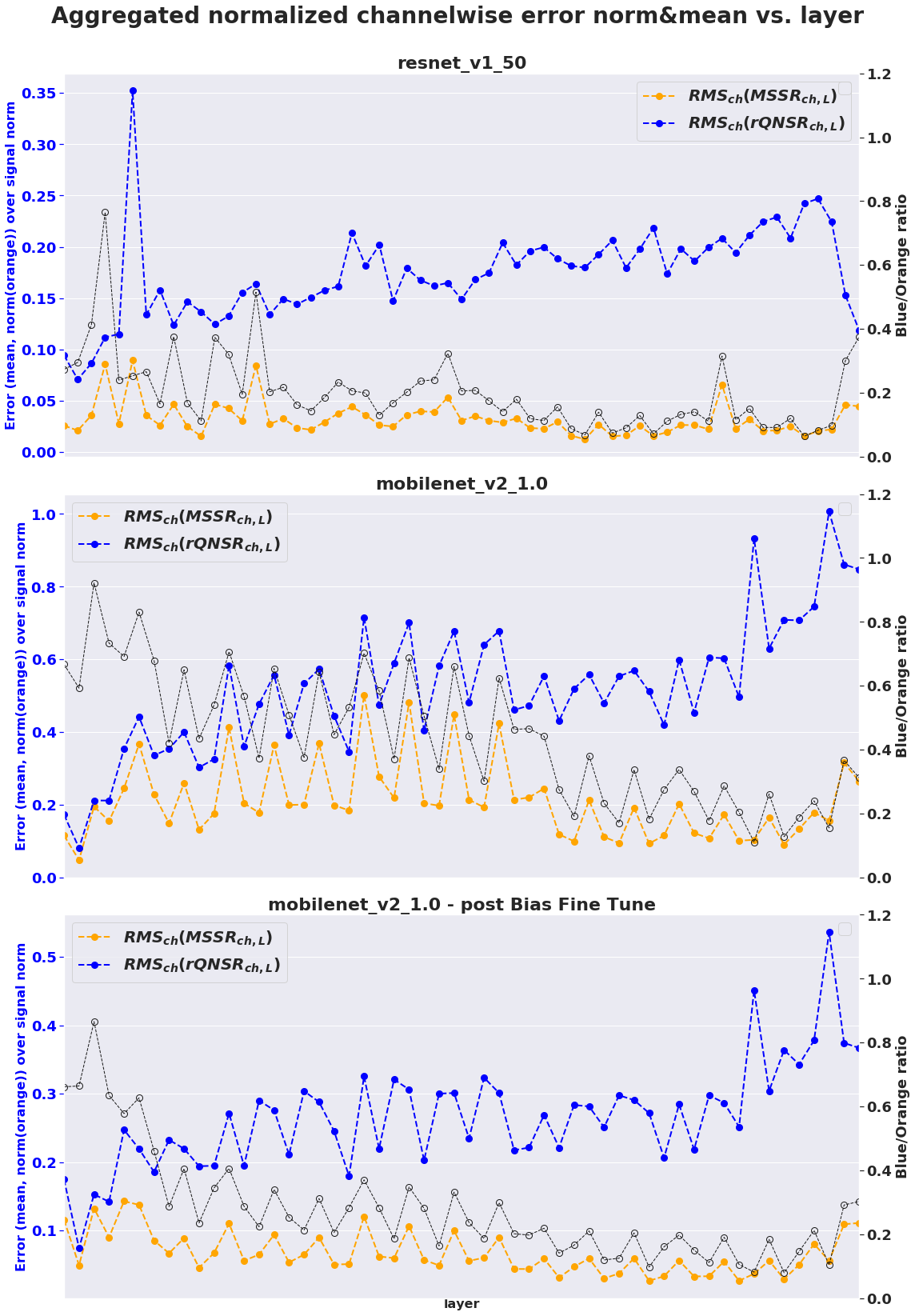}}			
			\caption{32-image estimates of the MSSR, QNSR, and their ratio ((a)/(b), (c)/(b), (a)/(c) ratios in terms of fig. (\ref{fig:MeanShiftContribution})) L2-aggregated (RMS) over channels, for all layers of same 2 nets. Comparing \textbf{top} and \textbf{center} we can see again, now at the whole net level, that for mobilenet the overall error is greater and it's driven in large part by the MAS. Some typical behaviors can be spotted - e.g. average-pooling layer reduces QNSR but the mean-shift contribution increases. Comparing \textbf{center} and \textbf{bottom} plots, the great error reduction following the BFT procedure is evident (IBC gives a similar result); being driven by reducing MAS, it's not surprising that the relative contribution of MAS (black line) goes down as well
			}
		\label{fig:MeanShiftContributionLayers}
		\end{center}
		\vskip -0.2in
\end{figure}

One view of MAS as defined in eq. \eqref{define_shift} is as the activations' distribution shift - between the original and the quantized net (see fig. \ref{fig:DistributionShift}). Another analysis avenue relates it to the MSE decomposition:
\begin{equation}\label{mean_var_decompose}
MSE = E(error^2) = {Mean^2}(error) + Var(error)
\end{equation}
In this framework, the question of the MAS's  significance can be formulated as gauging the \textbf{relative contribution} of the error mean to error magnitude, which in our notation can be written as: $\frac{E^2(e_{l, ch})}{E(e^2_{l, ch})}$. To answer this question we estimate these quantities on data samples obtained by running the full precision and the quantized nets over batches of few tens of images each to obtain signal and error vectors; see figs. \ref{fig:MeanShiftContribution}, \ref{fig:MeanShiftContributionLayers} (curves for different choices of batch are very similar; we drop all but one from plot for visual brevity). We can see that for many layers the contribution is significant, sometimes dominating the error energy. Fortunately, the degradation-complicit $\Delta_{l,ch}$ lends itself to a convenient correction, one that doesn't entail changing the NN's computational graph but only adjusting its parameters - namely, subtracting MAS from the (per-channel) biases $B_{l,ch}$ to be added pre-activation
\begin{footnote} {
 Referring to a generic convolutional layer with kernel of size $K$:
 $ X_{l,ch}=F_{act} \left(\sum_{ch',dx,dy}^{K} {X_{l-1,ch'}W_{l,ch}^{ch',dx,dy} + B_{l,ch}} \right) $
}
\end{footnote}
. For a single linear ($F_{act}=I$) layer, this bias-fixing operation can indeed be proven as the optimal error reduction by changing biases alone - effectively removing the Mean contribution in eq. \eqref{mean_var_decompose}. Note also that in contrast to the weights (and their potential adjustments), the bias is quantized at high precision in most implementations (the cost for this being rather low), so the above correction can be done relatively precisely.
 The natural mitigation taken together with the significance of the MAS contribution strikes a great cost-benefit balance, compared to e.g. quantization-aware training methods (and actually performing on-par with those in some cases as we shall see in section \ref{sec:Results}).  
 
Note that for a deep networks the situation is more complex since MAS for different channels in the first layers are mixed in deeper layers, causing input errors that may be comparable with (or larger than) the MAS of the deeper layers.

We will see how these issues are tackled by our optimization methods in section \ref{sec:Optimization Methods}.

\subsection{MAS Generation Mechanism in MobileNets}\label{subsection3.3}
We now wish to connect the number of weight elements in a layer and the typical magnitude of MSSR in that layer and show that small kernels tend to increase MAS. Let us consider layer $l$ in which the calculation of an output channel activation involves $k$ kernel elements and $k$ input activations. We define a \textbf{weight rounding error} as \textbf{$\delta_{W_i}$} where $i$ denotes the weight elements in the set of size $k$. 
The weight rounding errors are i.i.d and follow the distribution:
\begin{equation}\label{weight_dist}
f(\delta_{W_i}) = U[-\frac{max(|W|)}{2^{N-1}}, \frac{max(|W|)}{2^{N-1}}]
\end{equation}
The mean and standard-deviation of the sum of rounding errors is given by:

\begin{equation}\label{rounderrormean}
E(\sum^k(\delta_{W_i})) = 0
\end{equation}
\begin{equation}\label{rounderrorstd}
\sigma(\sum^k\delta_{W_i}) = C \sqrt{(k/12)}
\end{equation}
with $C$ depending only on $max(|W|)$ and $N$ where $max(|W|)$ is the maximum element magnitude in the weight kernel and $N$ is the number of bits used for quantized representation. 
It is worth noting that the assumption of a given $max(|W|)$ means that quantization grid is constant. The rounding error is the results of decimating the kernel values to this grid and so the expectation in the above equations is taken with respect to possible kernel values.

The activation shift for an input image, defined in \ref{define_error} can be expressed as:
\begin{equation}\label{mean_shift_full}
e_{l,ch} = \sum_{i=1}^k (x_{in}^i + \delta_{x_{in}}^i) \times (w_i + \delta_{w_i}) - \sum_{i=1}^k x_{in}^i \times w_i
\end{equation}
or,
\begin{equation}\label{mean_shift_full2}
e_{l,ch} = \sum_{i=1}^k \delta_{x_{in}}^i \times w_i + \sum_{i=1}^k x_{in}^i \times \delta_{w_i}
\end{equation}

In the first layer of the network, the first term in the right-hand side is negligible since when the input rounding errors are small such as in the first layer. We consider only the second term in the following analysis.
Averaging over input data set and the spatial dimensions of layer $l$ to achieve the numerator in \ref{define_MSSR}, we get:
\begin{equation}\label{Emean_shift_full2}
E(e_{l,ch}) = E(\sum_{i=1}^k x_{in}^i \times \delta_{w_i}) = E(x_{in}) \times \sum_{i=1}^k \delta_{w_i}
\end{equation}
where we made the assumption that the input data $x_i$ are i.i.d. The separation of terms in the mean is possible since for a given channel the $\delta_{w_i}$ are constant and independent of the data.

Let us rewrite $MSSR_{l,ch}$ as:
\begin{equation}\label{MSSR2}
MSSR_{l, ch} = \frac{E(x_{in})} {\sqrt{E(x_{out}^2)}} \times \sum_{i=1}^k \delta_{w_i}
\end{equation}
We regard the first term of the product as the 'data term'. If we treat the convolution as a matched filter it can be approximated as:
\begin{equation}\label{rms_prod}
\frac{E(x_{in})} {\sqrt{E(x_{out}^2)}} \approx 1/k
\end{equation}
Using the above approximation we express the mean and variance of $MSSR_{l,ch}$ with regards to possible kernel values in layer $l$ as:
\begin{equation}\label{MSSR2}
E(MSSR_{l, ch}) = C_1\times 1/k \times E(\sum_{i=1}^k\delta_{w_i}) = 0 
\end{equation}
\begin{align}\label{MSSR2}
\begin{split}
\sigma(MSSR_{l, ch}) = C_21\times 1/k  \times \sigma(\sum_{i=1}^k\delta_{w_i}) \\
\approx \sqrt{k}/k=1/\sqrt{k}
\end{split}
\end{align}

Equation \ref{MSSR2} tells us that layers with small kernels will tend to have larger MAS energy to activation energy ratio. 
This conclusion is in agreement with our observation that Mobilenet architecture exhibits strong mean activation shifts, since it incorporates many depthwise (dw) convolutions, involving small (9 elements) kernel slices. In contrast, conventional 2D convolutions typically involve hundreds of elements for the calculation of each output channel.

This is only one possible source of MAS and the occurrence of significant mean activation shift in Inception v1 despite the absence of depthwise layers suggests other possible sources for MAS. One such source can be the neglected term in \ref{mean_shift_full2} which becomes significant once the input of the layer exhibits large noise.

\section{Optimization Methods}\label{sec:Optimization Methods}
We now present two methods that compensate for the QNN MAS (sec. \ref{sec:Problem Statement}) by updating the bias terms. While the first (IBC) is very fast and requires a very small dataset, the second (BFT) is more general and utilizes a stronger optimizer.

\subsection{Iterative Bias Correction (IBC)}
Estimating the MAS of all the layers and channels and updating the bias terms of the network in one pass cannot be done, since correcting the bias terms of layer $l$ changes the input to layer $l + 1$, which, in turn, gives a different MAS than the one calculated before. Therefore, the correction process has to be performed \textbf{iteratively} according to algorithm \ref{IBC algorithm}, where $act^{orig}_{l, ch}$ and $act^{quant}_{l, ch}$ are the activation values of layer $l$, channel $ch$ in the original and quantized networks, respectively. The inner 'for' loop in the algorithm can be vectorized and done on all of the channels of a layer $l$ simultaneously. The set of input images used for the Iterative Bias Correction (short-named \textit{IBC-batch}), has a typical size of 8-64 images, and can be drawn from the same set used for the QNN calibration. We find that the Iterative Bias Correction effectiveness can vary with IBC-batch size, and larger batches do not necessarily produce better results. 
The strength of IBC lies in its simplicity and high speed. With as little as 8 unlabeled images (that can be just the quantization batch) and just a few minutes computation we were able to achieve the results in table \ref{results table}.

\begin{algorithm}
 \caption{Iterative Bias Correction (IBC). The inputs to the IBC algorithm are the original and quantized networks, a topologically sorted list of layers, where 0 is the input layer and the last layer is the output layer. In each iteration the output of the current layer for both the original and quantized networks is evaluated and averaged over $n$ images the spatial dimensions to give a single value per output channel $ch$. The MAS is computed and added to the layer bias.}
 \SetAlgoLined
 \KwResult{Corrected QNN checkpoint}
 evaluate $act^{orig}_{l, ch}$ for all $l$, $ch$ in original net\;
 $l \gets 0$\;
 \While{not end of original net}{
 evaluate $act^{quant}_{l, ch}$\;
 \For{$ch \in l$}{
 $\Delta_{l, ch} \gets <act^{orig}_{l, ch}> - <act^{quant}_{l, ch}> $\;
 $bias^{quant}_{l,ch} \gets bias^{quant}_{l,ch} + \Delta_{l, ch}$\;
 }
 $l \gets l+1$\;
 }
 \label{IBC algorithm}
\end{algorithm}

\subsection{Bias Fine Tuning (BFT)} \label{BFTdesc}
The observations discussed in section 3.2 and in fig. \ref{fig:MeanShiftContribution} suggest that modifications of the QNN's biases, reducing or otherwise modulating the MAS, have the potential to reduce noise energy, impact downstream QSNR and eventually the final layer's output accuracy. This line of thought suggests attempting a generic optimization of the biases alone, with the eventual objective of reducing the overall QNN's loss and accuracy degradation. To this end, we use a quantization-aware training procedure, which starts from the pre-trained weights and biases, but restricts the trainable variables to be the biases only. This restriction enables fast training on a minuscule part of the training set, as small as \textbf{1K} images, without significant overfit, since the number of trainable variables is several orders of magnitude less than with a full training. Our procedure uses global optmization in contrast to the local estimation procedure used in IBC. Drawing inspiration from \cite{Apprentice, PolinoDistillation}, we use pure teacher-student distillation loss, namely the cross-entropy between the logits of the full-precision and the quantized net. This enables a label-free training. To make the training quantization-aware, we use quantized weights for the trainable net, and add $FakeQuant$ operations of the \textit{TensorFlow} framework on the activations' path, implementing the standard Straight-Through Estimator (STE) approach \cite{Bengio_STE, understanding_STE}. Since the weights are fixed for the procedure, our method be also viewed as a miniature variant of Incremental Training \cite{zhou2017IntelINQincremental, zhuang2018GradualBitReduction}, in the sense that biases are optimized to compensate for the error generated by quantization of the weights. Finalizing our procedure, the fine-tuned full precision biases are re-quantized to 16 bits which is precise enough for this final step to have no accuracy impact. 

\section{Experiments}\label{sec:Results}
For our tests we used the following setup and procedure (all experiments were done with the \textit{TensorFlow} framework):
\begin{enumerate}
  \item Publicly available pre-trained full-precision models from \href{https://github.com/tensorflow/models/tree/master/research/slim}{tf-slim collection} is ingested; any BatchNorm ops are folded onto preceding layer's weights.
  \item For mobilenets: dead channels (channels with constant output / zero variance) are dropped.
  \item Weights are quantized onto a uniform, symmetric 8-bit (INT8), per-layer grid. Biases are quantized to 16-bit representation.
  \item Activations are quantized (at run-time) onto a fixed, uniform, asymmetric 8-bit (UINT8), per-layer grid over a range defined by simple $min,max$ statistics on a 64-image calibration batch.

  \item QNN accuracy is evaluated on the standard 50K ImageNet validation set. 
\end{enumerate}
Table 1 summarizes the results of simple-case tests over several neural nets. The second column gives the Top-1 accuracy of the original, floating-point representation net. The third column gives the top-1 accuracy degradation using the basic quantization described in \cite{Jacob2017}. The last 2 columns give the top-1 accuracy degradation of the quantized nets after applying the IBC and BFT optimization tools.

\begin{table}[]
 	\caption{Top-1 accuracy of original nets and accuracy degradation with basic 8-bit quantization, Iterative Bias Correction (IBC), and Bias Fine Tune (BFT). All numbers are in units of absolute percentage (\%). Best IBC results for were achieved with IBC batch size of 8.}
 	\begin{tabular}{|l||l||l|l|l|}\hline\label{results table}
 		\textbf{\begin{tabular}[c]{@{}l@{}}Network \\ Name\end{tabular}}&
 		\textbf{\begin{tabular}[c]{@{}l@{}} Original  \\ Top-1 \\ accuracy \end{tabular}}&
 		\textbf{\begin{tabular}[c]{@{}l@{}}Basic \\ Quant.\end{tabular}}& \textbf{\begin{tabular}[c]{@{}l@{}}IBC \end{tabular}}&
 		 \textbf{\begin{tabular}[c]{@{}l@{}}BFT\end{tabular}} \\
 		 \hline
 		Mobilenet-v1-1.0  &$71.02$& $7.90$   & $0.92$ & $1.03$ \\ \hline
 		Mobilenet-v2-1.0  &$71.8$& $16.44$   & $1.42$ & $1.2$ \\ \hline
 		Mobilenet-v2-1.4  &$74.95$& $6.42$   & $1.1$ & $0.85$ \\ \hline
 		Inception-v1      &$69.76$& $2.26$   & $0.44$ & $0.47$ \\ \hline
 		Resnet-50-v1      &$75.2$& $0.27$   & $>10$ & $0.30$ \\ \hline
  	\end{tabular}
\end{table}

\subsection{Experiments With IBC}
Testing IBC with different sizes of IBC batches between 8 to 64 images shows a variance of a few 0.1\% in results. Out of the cases we tested, the best results were achieved using an IBC batch of 8 images from the calibration batch.
While MobileNet and Inception architectures benefit from employing IBC, Resnets show large degradation by approximately 10\% when utilizing IBC. We were unable to understand the source of this degradation. 

The IBC algorithm discussed throughout this paper compares \textbf{post-activation} means in the original and quantized nets. In Mobilenet v1 the activation function is $Relu6$, clipping the output at 0 and 6, which results in a loss of information regarding the convolution output distribution. It would be reasonable to assume that updating the bias terms according to \textbf{pre-activation} distribution would utilize the full information available and produce even better results than those shown in table \ref{results table}. Indeed, testing a variant of algorithm \ref{IBC algorithm} in which $\Delta_{l, ch}$ is calculated \textbf{pre-activation} on Mobilenet v1, yields a degradation of 0.76\% -  $0.16\%$ better than the post-activation degradation given in table \ref{results table}). 

\subsection{Experiments With BFT}
 We trained using the standard Adam optimizer, with a learning rate schedule scanning a wide range as recommended in \cite{IBMFAQ2018discover4bit}: in our default schedule we use $10^{-3},10^{-4},10^{-5},10^{-6}$ rates for 16 mini-epochs each, for a total of 64 mini-epochs using the same 1K images. This schedule is used across all nets presented here (see table \ref{results table}), proving a high degree of robustness. That stands in contrast to the normally high appetite of training-based methods for expertise and data. Given the above, and especially the relaxed input requirement of only 1K label-less images, (similar in size to calibration sets typically used \cite{Migacz2017, KrishnaGoogle}), we claim that the procedure should be seen as a post-training quantization method despite sharing the "backprop+STE+SGD" approach with of quantization-aware training methods.

The results (table \ref{results table}) are quite similar to IBC (suggesting that both utilize well a common "resource" discussed above) - with the exception of ResNet which isn't degraded with BFT. On both Mobilenet-v1 and v2, we perform on par with the state-of-the-art of 1\% degradation reached with full quantization-aware training \cite{KrishnaGoogle}. The restriction to biases apparently enables harvesting the low-hanging fruits of quantization-aware training - which happen to include the lion's share of the degradation to be reduced - using a fraction of the cost.
When combining \textit{ChannelEqualization} \cite{SameSame} with BFT on the Mobilenet-v2 nets, we achieve state-of-the-art quantized net accuracy of 71.1\% (v2-1.0) and 74.3\% (v2-1.4).

\section{Discussion}
Mean activation shift (MAS) occurs in several neural net architectures after quantization. One progenitor for this shift was demonstrated to be a non-symmetrical distribution of the rounding error of weights, more pronounced when the weight kernels involved are small, such as in depthwise convolutions. The phenomenon was observed in Mobilenet v1 and v2 which incorporate depthwise convolutions. It was also observed in Inception v1 which incorporates 2D convolutions and concatenation layers, and no depthwise layers, leading to the conclusion that there are more sources for MAS.
  
We presented two methods that compensate for mean activation shift of nets by modifying biases - \textbf{Iterative Bias Correction (IBC)}, and \textbf{ Bias Fine Tuning (BFT)}.
Employing either one significantly improves the accuracy of Mobilenet v1, v2 and Inception v1, bringing their degradation down by an order of magnitudes relative to the baseline quantization. For Mobilenet-v1/v2, post-\textbf{BFT} degradation is on par with the ~1\% state-of-the-art \cite{KrishnaGoogle} achieved by resource-intensive quantization-aware training, tested with the same 8-bit quantization scheme. Both tools require very little in terms of tuning effort, run-time and data. BFT performs a robust micro-training procedure, incorporating a strong optimizer (gradient descent) and requiring $\httilde{1000}$ unlabeled images; typically taking $\httilde{20min}$ on a single GPU. IBC uses a simpler direct-estimation, runs even faster (2-3min is typical) and requires as little as 8 unlabeled images.
We expect these methods to readily extend to mobilenet-based nets used as a base for other tasks (detection, segmentation, etc.) and at other quantization schemes (e.g. 4-bit weights). Both may permit further improvement by using more data and tuning their parameters in a scheme- and net- specific way; however, we refrain from it here, emphasizing instead the "without-bells-and-whistles" effectiveness of the methods, which highlights our basic insight (sec. \ref{subsection3.2}) about the underlying issue. The default setting seems to be enough for sub-$1\%$ degradation on most presented nets (table \ref{results table}). On the other hand, in yet more challenging cases (e.g. quantization to $n<8$ bits) the prospect for the methods described is to be used as a part of a wider post-training quantization toolset. We leave that to future work.

{\small
\bibliographystyle{ieee}
\bibliography{egbib}

\begin{thebibliography}{10}\itemsep=-1pt

\bibitem{Soudry4bit}
R.~Banner, Y.~Nahshan, E.~Hoffer, and D.~Soudry.
\newblock Post training 4-bit quantization of convolution networks for
  rapid-deployment.
\newblock {\em CoRR}, abs/1810.05723, 2018.

\bibitem{baskin2018uniq}
C.~Baskin, E.~Schwartz, E.~Zheltonozhskii, N.~Liss, R.~Giryes, A.~M. Bronstein,
  and A.~Mendelson.
\newblock Uniq: uniform noise injection for the quantization of neural
  networks.
\newblock {\em arXiv preprint arXiv:1804.10969}, 2018.

\bibitem{Bengio_STE}
Y.~Bengio, N.~L{\'{e}}onard, and A.~C. Courville.
\newblock Estimating or propagating gradients through stochastic neurons for
  conditional computation.
\newblock {\em CoRR}, abs/1308.3432, 2013.

\bibitem{choi2018pact_ReluAlphaTrained}
J.~Choi, Z.~Wang, S.~Venkataramani, P.~I.-J. Chuang, V.~Srinivasan, and
  K.~Gopalakrishnan.
\newblock Pact: Parameterized clipping activation for quantized neural
  networks.
\newblock {\em arXiv preprint arXiv:1805.06085}, 2018.

\bibitem{choukrounHuawei2019low}
Y.~Choukroun, E.~Kravchik, and P.~Kisilev.
\newblock Low-bit quantization of neural networks for efficient inference.
\newblock {\em arXiv preprint arXiv:1902.06822}, 2019.

\bibitem{goncharenko2018fast}
A.~Goncharenko, A.~Denisov, S.~Alyamkin, and E.~Terentev.
\newblock Fast adjustable threshold for uniform neural network quantization.
\newblock {\em arXiv preprint arXiv:1812.07872}, 2018.

\bibitem{KrishnaGoogle}
R.~K. (Google).
\newblock Quantizing deep convolutional networks for efficient inference: A
  whitepaper.
\newblock {\em CoRR}, abs/1806.08342, 2018.

\bibitem{SongHanDeepCompression}
S.~Han, H.~Mao, and W.~J. Dally.
\newblock Deep compression: Compressing deep neural networks with pruning,
  trained quantization and huffman coding.
\newblock {\em arXiv preprint arXiv:1510.00149}, 2015.

\bibitem{hinton2015distilling}
G.~Hinton, O.~Vinyals, and J.~Dean.
\newblock Distilling the knowledge in a neural network.
\newblock {\em arXiv preprint arXiv:1503.02531}, 2015.

\bibitem{2017arXiv170404861H}
A.~G. {Howard}, M.~{Zhu}, B.~{Chen}, D.~{Kalenichenko}, W.~{Wang}, T.~{Weyand},
  M.~{Andreetto}, and H.~{Adam}.
\newblock {MobileNets: Efficient Convolutional Neural Networks for Mobile
  Vision Applications}.
\newblock {\em arXiv e-prints}, page arXiv:1704.04861, Apr 2017.

\bibitem{Huang2016}
G.~Huang, Z.~Liu, and K.~Q. Weinberger.
\newblock Densely connected convolutional networks.
\newblock {\em CoRR}, abs/1608.06993, 2016.

\bibitem{Jacob2017}
B.~Jacob, S.~Kligys, B.~Chen, M.~Zhu, M.~Tang, A.~Howard, H.~Adam, and
  D.~Kalenichenko.
\newblock {Quantization and Training of Neural Networks for Efficient
  Integer-Arithmetic-Only Inference}.
\newblock 2017.

\bibitem{Jung2018}
S.~Jung, C.~Son, S.~Lee, J.~Son, Y.~Kwak, J.~Han, and C.~Choi.
\newblock Joint training of low-precision neural network with quantization
  interval parameters.
\newblock {\em CoRR}, abs/1808.05779, 2018.

\bibitem{IBMFAQ2018discover4bit}
J.~L. McKinstry, S.~K. Esser, R.~Appuswamy, D.~Bablani, J.~V. Arthur, I.~B.
  Yildiz, and D.~S. Modha.
\newblock Discovering low-precision networks close to full-precision networks
  for efficient embedded inference.
\newblock {\em arXiv preprint arXiv:1809.04191}, 2018.

\bibitem{SameSame}
E.~Meller, A.~Finkelstein, U.~Almog, and M.~Grobman.
\newblock Same, same but different - recovering neural network quantization
  error through weight factorization.
\newblock {\em CoRR}, abs/1902.01917, 2019.

\bibitem{Migacz2017}
S.~Migacz.
\newblock {8-bit Inference with TensorRT}.
\newblock 2017.

\bibitem{Apprentice}
A.~K. Mishra and D.~Marr.
\newblock Apprentice: Using knowledge distillation techniques to improve
  low-precision network accuracy.
\newblock {\em CoRR}, abs/1711.05852, 2017.

\bibitem{park2018value_mixedprec}
E.~Park, S.~Yoo, and P.~Vajda.
\newblock Value-aware quantization for training and inference of neural
  networks.
\newblock In {\em Proceedings of the European Conference on Computer Vision
  (ECCV)}, pages 580--595, 2018.

\bibitem{PolinoDistillation}
A.~Polino, R.~Pascanu, and D.~Alistarh.
\newblock Model compression via distillation and quantization.
\newblock {\em CoRR}, abs/1802.05668, 2018.

\bibitem{2018arXiv180104381S}
M.~{Sandler}, A.~{Howard}, M.~{Zhu}, A.~{Zhmoginov}, and L.-C. {Chen}.
\newblock {MobileNetV2: Inverted Residuals and Linear Bottlenecks}.
\newblock {\em arXiv e-prints}, page arXiv:1801.04381, Jan 2018.

\bibitem{xilinxSmallFloat}
S.~O. Settle, M.~Bollavaram, P.~D'Alberto, E.~Delaye, O.~Fernandez, N.~Fraser,
  A.~Ng, A.~Sirasao, and M.~Wu.
\newblock Quantizing convolutional neural networks for low-power
  high-throughput inference engines.
\newblock {\em arXiv preprint arXiv:1805.07941}, 2018.

\bibitem{QuallcomMobilenet}
T.~Sheng, C.~Feng, S.~Zhuo, X.~Zhang, L.~Shen, and
  M.~Alf2018arXiv180104381Seksic.
\newblock A quantization-friendly separable convolution for mobilenets.
\newblock {\em CoRR}, abs/1803.08607, 2018.

\bibitem{Szegedy2014}
C.~Szegedy, W.~Liu, Y.~Jia, P.~Sermanet, S.~E. Reed, D.~Anguelov, D.~Erhan,
  V.~Vanhoucke, and A.~Rabinovich.
\newblock Going deeper with convolutions.
\newblock {\em CoRR}, abs/1409.4842.

\bibitem{PeteWarden17}
P.~Warder.
\newblock {What I’ve learned about neural network quantization}.
\newblock {\em petewarden.com}, 2008.

\bibitem{understanding_STE}
P.~Yin, J.~Lyu, S.~Zhang, S.~J. Osher, Y.~Qi, and J.~Xin.
\newblock Understanding straight-through estimator in training activation
  quantized neural nets.
\newblock In {\em International Conference on Learning Representations}, 2019.

\bibitem{Zhang2017}
X.~Zhang, X.~Zhou, M.~Lin, and J.~Sun.
\newblock Shufflenet: An extremely efficient convolutional neural network for
  mobile devices.
\newblock {\em CoRR}, abs/1707.01083, 2017.

\bibitem{zhao2019ChannelSplitting}
R.~Zhao, Y.~Hu, J.~Dotzel, C.~De~Sa, and Z.~Zhang.
\newblock Improving neural network quantization using outlier channel
  splitting.
\newblock {\em arXiv preprint arXiv:1901.09504}, 2019.

\bibitem{zhou2017IntelINQincremental}
A.~Zhou, A.~Yao, Y.~Guo, L.~Xu, and Y.~Chen.
\newblock Incremental network quantization: Towards lossless cnns with
  low-precision weights.
\newblock {\em arXiv preprint arXiv:1702.03044}, 2017.

\bibitem{zhuang2018GradualBitReduction}
B.~Zhuang, C.~Shen, M.~Tan, L.~Liu, and I.~Reid.
\newblock Towards effective low-bitwidth convolutional neural networks.
\newblock In {\em Proceedings of the IEEE Conference on Computer Vision and
  Pattern Recognition}, pages 7920--7928, 2018.

\end{thebibliography}
}

\end{document}